\documentclass{article}

\usepackage{microtype}
\usepackage{graphicx}
\usepackage{caption,subcaption}
\usepackage{booktabs} %
\usepackage{amsfonts,amsmath,amssymb,amsthm}
\usepackage{todonotes}

\allowdisplaybreaks

\usepackage[accepted]{icml2021}

\icmltitlerunning{Leveraging Hidden Structure in Self-Supervised Learning}

\begin{document}

\twocolumn[
\icmltitle{Leveraging Hidden Structure in Self-Supervised Learning}

\icmlsetsymbol{equal}{*}

\begin{icmlauthorlist}
\icmlauthor{Emanuele Sansone}{to}
\end{icmlauthorlist}

\icmlaffiliation{to}{Department of Computer Science, KU Leuven, Belgium}

\icmlcorrespondingauthor{Emanuele Sansone}{emanuele.sansone@kuleuven.be}

\icmlkeywords{Representation Learning, Graph Neural Networks, Mutual Information, Self-Supervised Learning}

\vskip 0.3in
]

\printAffiliationsAndNotice{}  %

\begin{abstract}
This work considers the problem of learning structured representations from raw images using self-supervised learning. We propose a principled framework based on a mutual information objective, which integrates self-supervised and structure learning. Furthermore, we devise a post-hoc procedure to interpret the meaning of the learnt representations. Preliminary experiments on CIFAR-10 show that the proposed framework achieves higher generalization performance in downstream classification tasks and provides more interpretable representations compared to the ones learnt through traditional self-supervised learning.
\end{abstract}

{\section{Introduction}\label{sec:intro}}
Self-supervised learning has gained popularity in the last years thanks to the remarkable successes achieved in many areas, including natural language processing and computer vision. In this work, we consider extending self-supervised learning to learn structured representations directly from still images.

The contributions consist of (i) a new self-supervised framework, which leverages the hidden structure contained in the data and (ii) a learning procedure to interpret the meaning of the learnt representations.

The paper starts with the description of the proposed framework in Section~\ref{sec:theory}, continues with the discussion of the related work in Section~\ref{sec:related} and concludes with the experiments in Section~\ref{sec:experiments}.

{\section{Proposed Framework}\label{sec:theory}}
\begin{figure*}[t]
\vskip 0.2in
\begin{center}
     \begin{subfigure}[b]{0.41\textwidth}
         \centering
         \includegraphics[width=\textwidth]{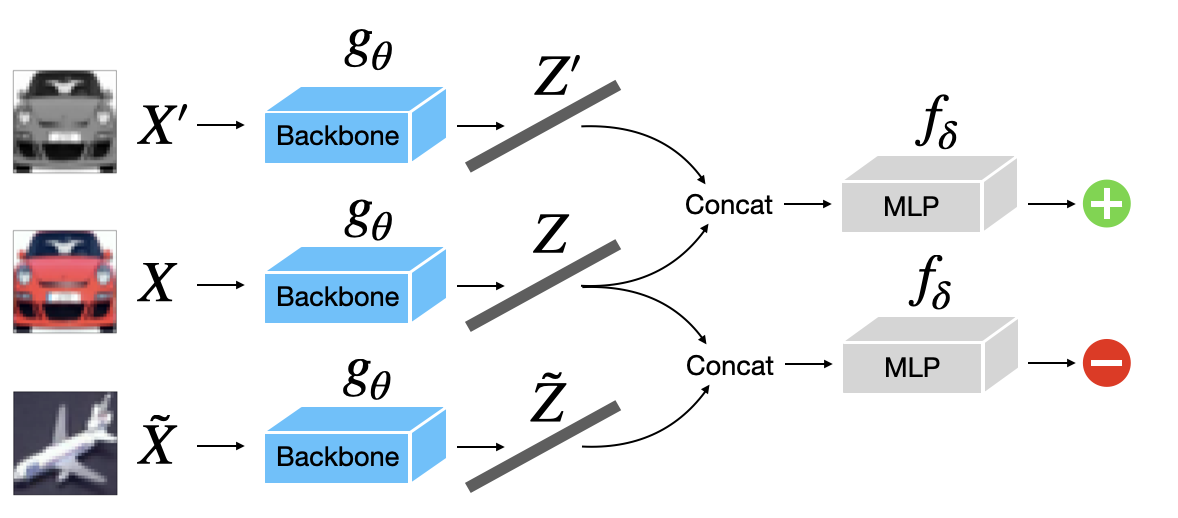}
         \caption{$T_{\pmb{\beta}}(\cdot,\cdot)$}
     \end{subfigure}
     \hfill
     \begin{subfigure}[b]{0.56\textwidth}
         \centering
         \includegraphics[width=\textwidth]{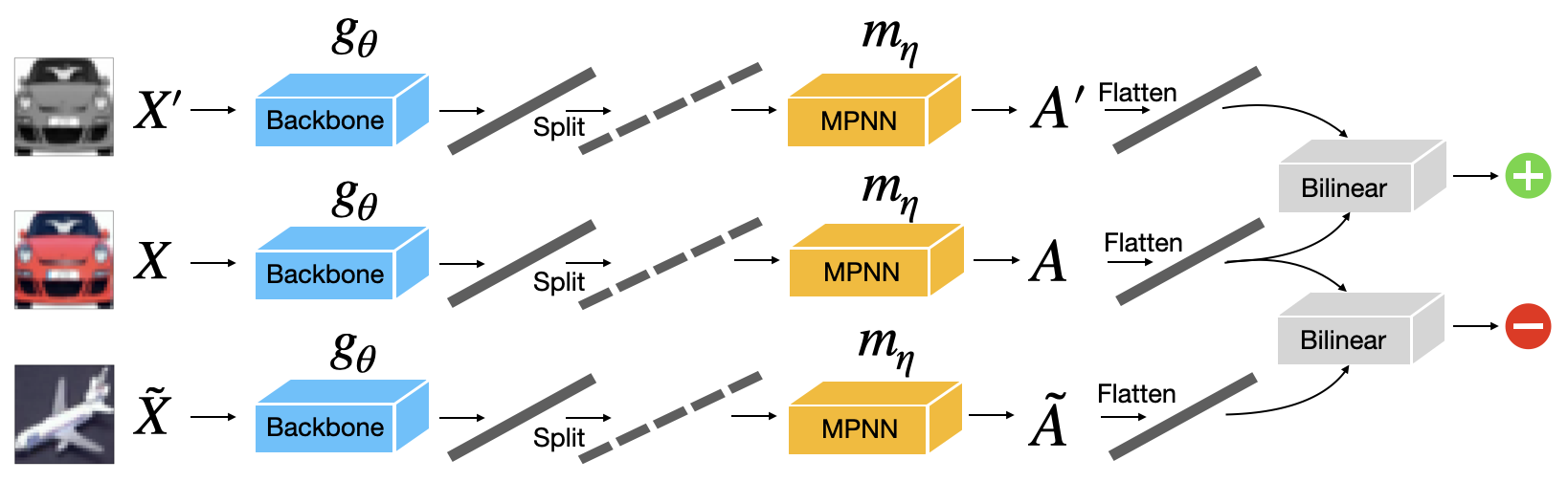}
         \caption{$T_{\pmb{\omega}}(\cdot,\cdot)$}
     \end{subfigure}
     \caption{Visualization of the two critic functions in Eq.~\ref{eq:bound}. Note that the critic functions $T_{\pmb{\beta}}$ and $T_{\pmb{\omega}}$ are applied to pairs of positive samples, i.e. $\pmb{x}$ and its augmented version $\pmb{x}'$, and pairs of negative samples, i.e. $\pmb{x}$ and $\tilde{\pmb{x}}$. Positive and negative pairs are used to estimate the first and the last two addends of Eq.~\ref{eq:bound}, respectively. Regarding the network architectures, we used a Conv4 and a Resnet-8 for $g_{\pmb{\theta}}$ and a multi-layer perceptron network for $f_{\pmb{\delta}}$, using the settings of~\cite{patacchiola2020self}. For $m_{\pmb{\eta}}$, we used a two-layer message passing neural network (MPNN) with a complete graph, following~\cite{kipf2018neural} (more details are provided in the Appendix).}
     \label{fig:model}
\end{center}
\vskip -0.2in
\end{figure*}

\begin{figure}[t]
\vskip 0.2in
\begin{center}
    \includegraphics[width=0.8\columnwidth]{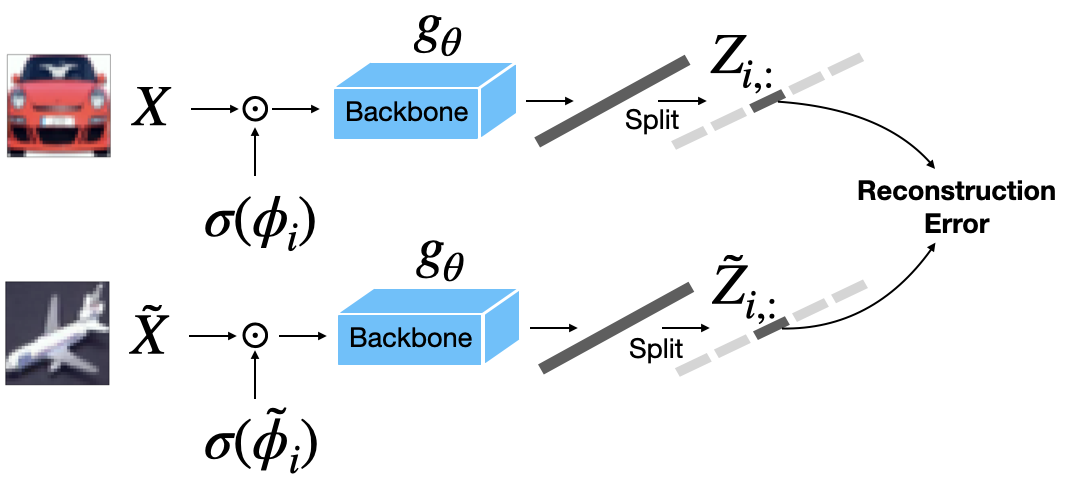}
    \caption{Strategy used to analyze the representations learnt by the encoder.}
    \label{fig:analysis}
\end{center}
\vskip -0.2in
\end{figure}

\subsection{Objective}
Let us consider an input tensor $\pmb{x}\in\mathbb{R}^{H\times W\times C}$, a latent tensor $\pmb{z}\in\mathbb{R}^{S\times D}$, obtained by the transformation of $\pmb{x}$ using an encoding function $g_{\pmb{\theta}}:\mathbb{R}^{H\times W\times C}\rightarrow\mathbb{R}^{S\cdot D}$ and a reshaping operator, and a latent structure tensor $\pmb{a}\in \{0,1\}^{S\times S\times K}$, obtained by the transformation of $\pmb{x}$ using function $h_{\pmb{\gamma}}:\mathbb{R}^{H\times W\times C}\rightarrow \{0,1\}^{S\times S\times K}$. For example, $\pmb{x}$ can be an observed image, $\pmb{z}$ contains the representations of $S$ entities in the image and $\pmb{a}$ represents the relations among these $S$ entities.\footnote{$K$ stands for the relation type, including the case of no relation between two entities. We experimented only with $K=2$.}
If we define the random vectors/tensors $X,Z,A$, associated to $\pmb{x},\pmb{z}$ and $\pmb{a}$, respectively, we can define the generalized mutual information over these random quantities (equivalently the total correlation of $X,Z,A$):
\begin{align}
    \mathcal{I}(X,Z,A)\doteq KL\big\{p_{X,Z,A}||p_X p_Z p_A\big\}
    \label{eq:total}
\end{align}
where $KL$ is the Kullback-Leibler divergence and $p_{X,Z,A}$, $p_X, p_Z, p_A$ are the joint probability and the marginal densities of the random quantities, respectively.

Now, let us assume that $Z$ and $A$ are conditionally independent given $X$. Note that this is quite reasonable in practice. For example, if we consider an image containing a set of objects and a set of relations (e.g. right/left of, above/beyond, smaller/bigger), these two sets exist independently from each other and are instantiated through $\pmb{z}$ and $\pmb{a}$ by the observed scene. Thanks to this assumption, Eq.~\ref{eq:total} simplifies into the following form (see the Appendix for the derivation):
\begin{align}
    \mathcal{I}(X,Z,A)=\mathcal{I}(X,Z)+\mathcal{I}(X,A)
    \label{eq:sum}
\end{align}
Now, we are ready to provide an estimator for the quantity in Eq.~\ref{eq:sum}, by using existing variational bounds~\cite{agakov2004algorithm}. Specifically, we focus on the bound proposed in~\cite{nguyen2010estimating} for its tractability and unbiasedness, thus obtaining the following criterion (see the Appendix for the derivation):
\begin{align}
    \mathcal{I}(X,Z,A) &\geq E_{p_{X,Z}}\{T_{\pmb{\beta}}(\pmb{x},\pmb{z})\} + E_{p_{X,A}}\{T_{\pmb{\omega}}(\pmb{x},\pmb{a})\} \nonumber\\
    &\quad{-}\frac{1}{e}\Big[ E_{p_X p_Z}\{e^{T_{\pmb{\beta}}(\pmb{x},\pmb{z})}\}{+}
    E_{p_X p_A}\{e^{T_{\pmb{\omega}}(\pmb{x},\pmb{a})}\}
    \Big]
    \label{eq:bound}
\end{align}
where $T_{\pmb{\beta}}(\pmb{x},\pmb{z})$ and $T_{\pmb{\omega}}(\pmb{x},\pmb{a})$ are two critic functions. Specifically, we have that $T_{\pmb{\beta}}(\pmb{x},\pmb{z})\doteq f_{\pmb{\delta}}(g_{\pmb{\theta}}(\pmb{x}),\pmb{z})$, where  $f_{\pmb{\delta}}:\mathbb{R}^{S\times D}\rightarrow\mathbb{R}$ and $\pmb{\beta}\doteq[{\pmb{\theta}},{\pmb{\delta}}]$ are the concatenated parameters of $T_{\pmb{\beta}}(\pmb{x},\pmb{z})$. Furthermore, we have that $T_{\pmb{\omega}}(\pmb{x},\pmb{a})\doteq\sum_{k=1}^K\text{Flatten}(h_{\pmb{\gamma}}(\pmb{x})_{:,:,k})^T \pmb{w}_{:,:,k}\text{Flatten}(\pmb{a}_{:,:,k})$, where $\pmb{w}\in\mathbb{R}^{S^2\times S^2\times K}$, $h_{\pmb{\gamma}}\doteq m_{\pmb{\eta}}\circ g_{\pmb{\theta}}$, $m_{{\pmb{\eta}}}:\mathbb{R}^{S\times D}\rightarrow\{0,1\}^{S\times S\times K}$, and $\pmb{\omega}=[{\pmb{\theta}},{\pmb{\eta}},\pmb{w}]$ are the concatenated parameters of $T_{\pmb{\omega}}(\pmb{x},\pmb{a})$.
For simplicity, we visualize and describe all these terms, including the two critic functions, in Figure~\ref{fig:model}.

Therefore, the learning problem consists of a maximization of the lower bound in Eq.~\ref{eq:bound} with respect to parameters $\pmb{\beta}$ and $\pmb{\omega}$. During training the encoder learns to structure the representation guided by the two critic functions to improve not only the mutual information between $X$ and $Z$ but also the mutual information between $X$ and $A$. In effect, this is an inductive bias towards learning more structured representations.

\subsection{Interpreting the Learnt Representations}
From Figure~\ref{fig:model} we can see that the encoder $g_{\pmb{\theta}}$ provides a global representation of the data, which is subsequently splitted/reshaped to provide the latent tensor $\pmb{z}$. It's interesting to understand the role of the rows of $\pmb{z}$ once the network is trained, to test (i) whether the network learns to bind each row of $\pmb{z}$ to a particular portion of the input and (ii) whether such portion corresponds to a particular entity of the scene.

To do so, we need to identify which entries of the input tensor (i.e. which pixels) mostly affect the different rows of $\pmb{z}$. Specifically, we introduce a soft-binary mask $\sigma(\pmb{\phi}_i)$ for each input sample and each row $i$ of $\pmb{z}$, where $\sigma$ is an element-wise standard logistic function and $\pmb{\phi}_i\in\mathbb{R}^{H\times W}$ is the parameter tensor of the mask associated to a specific image and representation $\pmb{z}_{i,:}$, and then multiply the mask with the associated input. The final problem consists of minimizing $\mathcal{I}(X,Z_{i,:})$ for $i\in\{1,\dots,S\}$ with respect to the parameters of the mask $\pmb{\phi}_i$.

Similarly to previous case, we need to consider a tractable estimator for the mutual information objective. Differently from previous case, we devise an upper bound to be consistent with the minimization problem (see the Appendix for the whole derivation), namely:
\begin{align}
 \mathcal{I}(X,Z_{i,:}) &\leq E_{p_{X,Z_{i,:}}}\Big\{\log\frac{p_{Z_{i,:}|X}}{q_{Z_{i,:}}}\Big\} \nonumber\\
 &=E_{p_X p_{Z_{i,:}}}\Big\{\|\tilde{\pmb{z}}_{i,:}-g_{\pmb{\theta}}(\sigma(\pmb{\phi}_i)\odot\pmb{x})\|_2^2\Big\} + c
 \label{eq:upper}
\end{align}
where $q_{Z_{i,:}}\doteq\frac{E_{p_X}\{\mathcal{N}(\tilde{\pmb{z}}_{i,:}|g_{\pmb{\theta}}(\sigma(\pmb{\phi}_i)\odot\pmb{x}),I)\}}{\mathcal{Z}}$ is an auxiliary distribution defined over the latent tensor representation, $\mathcal{N}(\tilde{\pmb{z}}_{i,:}|g_{\pmb{\theta}}(\sigma(\pmb{\phi}_i)\odot\pmb{x}),I)$ is a multivariate Gaussian density with mean $g_{\pmb{\theta}}$ and identity covariance matrix and $c$ is a constant with respect to mask parameters. The bound in Eq.~\ref{eq:upper} becomes tight when $q_{Z_{i,:}}$ matches the true marginal $p_{Z_{i,:}}$. It's also interesting to mention that, in theory, the bound in Eq.~\ref{eq:upper} admits a trivial solution with all masks set to zero. However, in practice, we didn't observe the convergence to such trivial solutions, probably due the highly nonlinearity of the minimization problem.

Figure~\ref{fig:analysis} shows the building block of the minimization procedure. Once the strategy concludes, we can inspect the latent representations of the encoder by plotting the masks $\sigma(\pmb{\phi}_i)$ for each image and each latent vector $\pmb{z}_{i,:}$, as discussed in the experimental section.

{\section{Related Work}\label{sec:related}}
We organize the existing literature in terms of objectives, network architectures and data augmentations/pretext tasks.

\textbf{Objectives.} Mutual information is one of the most common criteria used in self-supervised learning~\cite{linsker1988self,becker1992self}. Estimating and optimizing mutual information from samples is notoriously difficult~\cite{mcallester2020formal}, especially when dealing with high-dimensional data. Most recent approaches focus on devising variational lower bounds on mutual information~\cite{agakov2004algorithm}, thus casting the problem of representation learning as a maximization of these lower bounds. Indeed, several popular objectives, like the mutual information neural estimation (MINE)~\cite{belghazi2018mutual}, deep InfoMAX~\cite{hjelm2018learning}, noise contrastive estimation (InfoNCE)~\cite{henaff2020data} to name a few,  all belong to the family of variational bounds~\cite{poole2019variational}. All of these estimators have different properties in terms of bias-variance trade-off~\cite{tschannen2019mutual,song2020understanding}. Our work generalizes over previous objectives based on mutual information, by considering structure as an additional latent variable. Indeed, we focus on learning a structured representation, which aims at better supporting downstream reasoning tasks and improving its interpretability. Therefore, our work is orthogonal to these previous studies, as different estimators can be potentially employed for this extended definition of mutual information. %

It's important to mention that new self-supervised objectives have been emerged recently~\cite{zbontar2021barlow,grill2020bootstrap} as a way to avoid using negative samples, which are typically required by variational bound estimators. However, these objectives consider to learn distributed representations and disregard the notion of hidden structure, which is currently analyzed in this work.

\textbf{Architectures.} Graph neural networks are one of the most common architectures used in self-supervised learning on relational data~\cite{kipf2016variational,kipf2018neural,davidson2018hyperspherical,velickovic2019deep}. While these works provide representations of the underlying graph data, thus supporting downstream tasks like graph/node classification and link prediction/graph completion, they typically assume that node/entities/symbols are known a priori. This work relaxes this assumption and devises a criterion able to integrate traditional representation learning with graph representation learning in a principled way.

Autoencoding architectures are also widely spread in representation learning. However, traditional VAE-based architectures~\cite{kingma2014auto,rezende2014stochastic} are typically not enough to learn disentangled representations~\cite{locatello2019challenging}. Recent work on autoencoders focuses on introducing structured priors by learning both object representations~\cite{greff2017neural} and their relations~\cite{van2018relational,goyal2021recurrent,stanic2020hierarchical} directly from raw perceptual data in dynamic environments. Autoencoders have been also applied to synthetic cases to perform unsupervised scene decomposition from still images~\cite{engelcke2019genesis,engelcke2021genesis,locatello2020object}, thus providing representations disentangling objects in a better way. Commonly to these works, we aim at learning object representations together with their relational information. Differently from these works, we compute an objective criterion, which does not require using a reconstruction term at the pixel level (thus avoiding learning noisy regularities from the raw data) and also avoids using a decoder, thus increasing the computational efficiency.

\textbf{Data augmentations/Pretext tasks}.%
Data augmentation strategies are typically used to produce transformed versions of data, which are then used to define positive and negative instances~\cite{hjelm2018learning,henaff2020data,patacchiola2020self}. Different views~\cite{bachman2019learning,tian2019contrastive} or different data modalities~\cite{gu2020self} can be also exploited to augment
the learning problem. Furthermore, there is a huge amount of related work devoted to the design of pretext tasks, namely predicting augmentations at the patch level~\cite{dosovitskiy2014discriminative}, predicting relative location of patches~\cite{doersch2015unsupervised}, generating missing patches~\cite{pathak2016context}, solving jigsaw puzzles~\cite{noroozi2016unsupervised}, learning to count~\cite{noroozi2017representation}, learning to spot artifacts~\cite{jenni2018self}, predicting image rotations~\cite{komodakis2018unsupervised}, predicting information among different channels~\cite{zhang2017split} and predicting color information from gray images~\cite{zhang2016colorful,larsson2016learning}. All of these strategies are complementary to our work. However, we believe that our generalized objective can enable the development of new pretext tasks leveraging also relational information.

{\section{Experiments}\label{sec:experiments}}
\begin{figure}[t]
\vskip 0.2in
\begin{center}
     \begin{subfigure}[b]{0.23\textwidth}
         \centering
         \includegraphics[width=\textwidth]{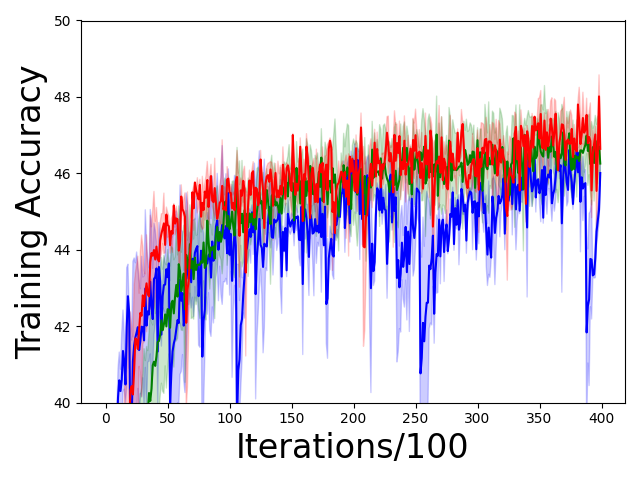}
         \caption{Conv-4}
     \end{subfigure}
     \hfill
     \begin{subfigure}[b]{0.23\textwidth}
         \centering
         \includegraphics[width=\textwidth]{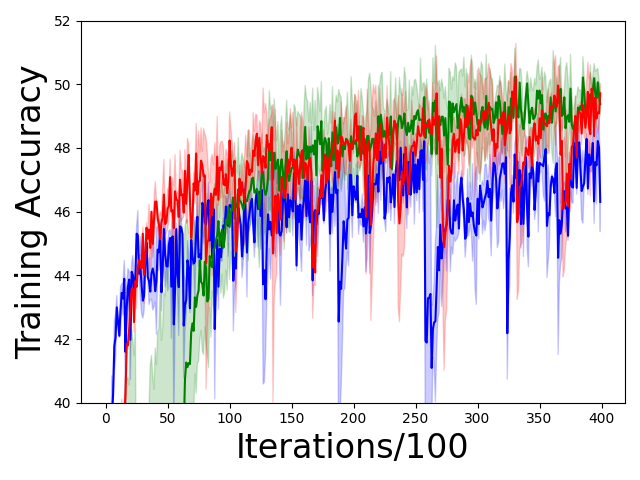}
         \caption{Resnet-8}
     \end{subfigure}
     \caption{Training accuracy obtained by a downstream linear classifier on CIFAR-10, during the unsupervised training of the backbones (Conv-4 on the left and Resnet-8 on the right). We trained only the linear classifier for $1$ epoch every $100$ unsupervised iterations. Blue, green and red curves correspond to the performance obtained by the method in~\cite{patacchiola2020self}, i.e. $\mathcal{I}(X,Z)$, our method with only relational information, i.e. $\mathcal{I}(X,A)$, and our complete method, i.e. $\mathcal{I}(X,Z,A)$, respectively.}
     \label{fig:accuracy}
\end{center}
\vskip -0.2in
\end{figure}

\begin{table}[t]
\caption{Test accuracies for linear evaluation with different backbones, viz. Conv-4 and Resnet-8. Self-supervised training on CIFAR-10 for $50$ epochs and supervised linear training on CIFAR-10 and CIFAR-100 for $100$ epochs.}
\label{tab:results}
\vskip 0.15in
\tiny
\begin{center}
\begin{small}
\begin{sc}
\resizebox{\columnwidth}{!}{
\begin{tabular}{l|cc|cc}
\toprule
& \multicolumn{2}{c|}{CIFAR-10} & \multicolumn{2}{c}{CIFAR-10$\rightarrow$CIFAR-100} \\
Method & Conv-4 & Resnet-8 & Conv-4 & Resnet-8 \\
\midrule
Supervised & 78.7$\pm$0.1 & 86.0$\pm$0.4 & 31.3$\pm$0.2 & 36.8$\pm$0.3 \\
Random & 33.2$\pm$1.9 & 36.8$\pm$0.2 & 11.0$\pm$0.4 & 13.6$\pm$0.3 \\
\midrule
$\mathcal{I}(X,Z)$* & 54.2$\pm$0.1 & 56.9$\pm$0.4 & 27.3$\pm$0.3 & 28.2$\pm$0.5 \\
$\mathcal{I}(X,A)$ (ours) & 55.9$\pm$0.8 & \pmb{60.0}$\pm$\pmb{0.1} & 26.2$\pm$0.8 & 28.4$\pm$0.1 \\
$\mathcal{I}(X,A,Z)$ (ours) & \pmb{56.3}$\pm$\pmb{0.8} & 59.5$\pm$0.2 & \pmb{27.8}$\pm$\pmb{0.4} & \pmb{29.6}$\pm$\pmb{0.5} \\
\midrule
\multicolumn{5}{l}{*\cite{patacchiola2020self}} \\
\bottomrule
\end{tabular}}
\end{sc}
\end{small}
\end{center}
\vskip -0.1in
\end{table}

The experiments are organized in a way to answer the following three research questions:
\begin{itemize}
    \item \textbf{Q1 (Surrogate)}: Is the proposed objective a good unsupervised surrogate for traditional supervised learning?
    \item \textbf{Q2 (Generalization)}: How well does the learnt representations generalize to downstream supervised tasks? And how well do they transfer?
    \item \textbf{Q3 (Interpretation)}: Can we interpret what the network has learnt to represent?
\end{itemize}

Given the limited amount of available GPU resources,\footnote{All experiments are run on 4 GPUs NVIDIA GeForce GTX 1080 Ti.} we run experiments on small backbones, namely Conv-4 and Resnet-8, and train them in an unsupervised way on CIFAR-10~\cite{krizhevsky2009learning} for $50$ epochs based on the code and the settings of~\cite{patacchiola2020self} (additional details are available in the Appendix). All experiments are averaged over three repeated trials. As a baseline, we consider the recent self-supervised framework in~\cite{patacchiola2020self}, which does not leverage structure and thus maximizes $\mathcal{I}(X,Z)$. For our framework, we consider two case, namely (i) the one using only structure, i.e. maximizing $\mathcal{I}(X,A)$, and (ii) the overall one, i.e. maximizing $\mathcal{I}(X,Z,A)$.

\textbf{Surrogate.} We evaluate the downstream training performance of a linear classifier during the unsupervised training of the backbone. Specifically, we train a classifier on the training set of CIFAR-10 for $1$ epoch every $100$ unsupervised iterations and report its training accuracy.  Figure~\ref{fig:accuracy} provides the evolution of the training accuracies for the two backbones. From the figures, we see that the baseline performs well during the early stages of training, but it is significantly outperformed by our purely structured-based strategy in the subsequent stages. Furthermore, the overall strategy is able to combine the positive aspects of the other strategies, thus providing a more principled approach.

\textbf{Generalization.} We evaluate generalization (CIFAR-10) and transfer (CIFAR-100) performance using a linear classifier. In this set of experiments, we also include two additional baselines, namely (i) a random one, i.e. an encoder initialized with random weights, and (ii) a supervised one, i.e. an encoder trained in a supervised fashion. Table~\ref{tab:results} summarizes the performance of all strategies. From these results, we see that our proposed framework allows for significant improvement both in terms of generalization and transfer over the baselines.

\begin{figure}[t]
\vskip 0.2in
\begin{center}
    \includegraphics[width=\columnwidth]{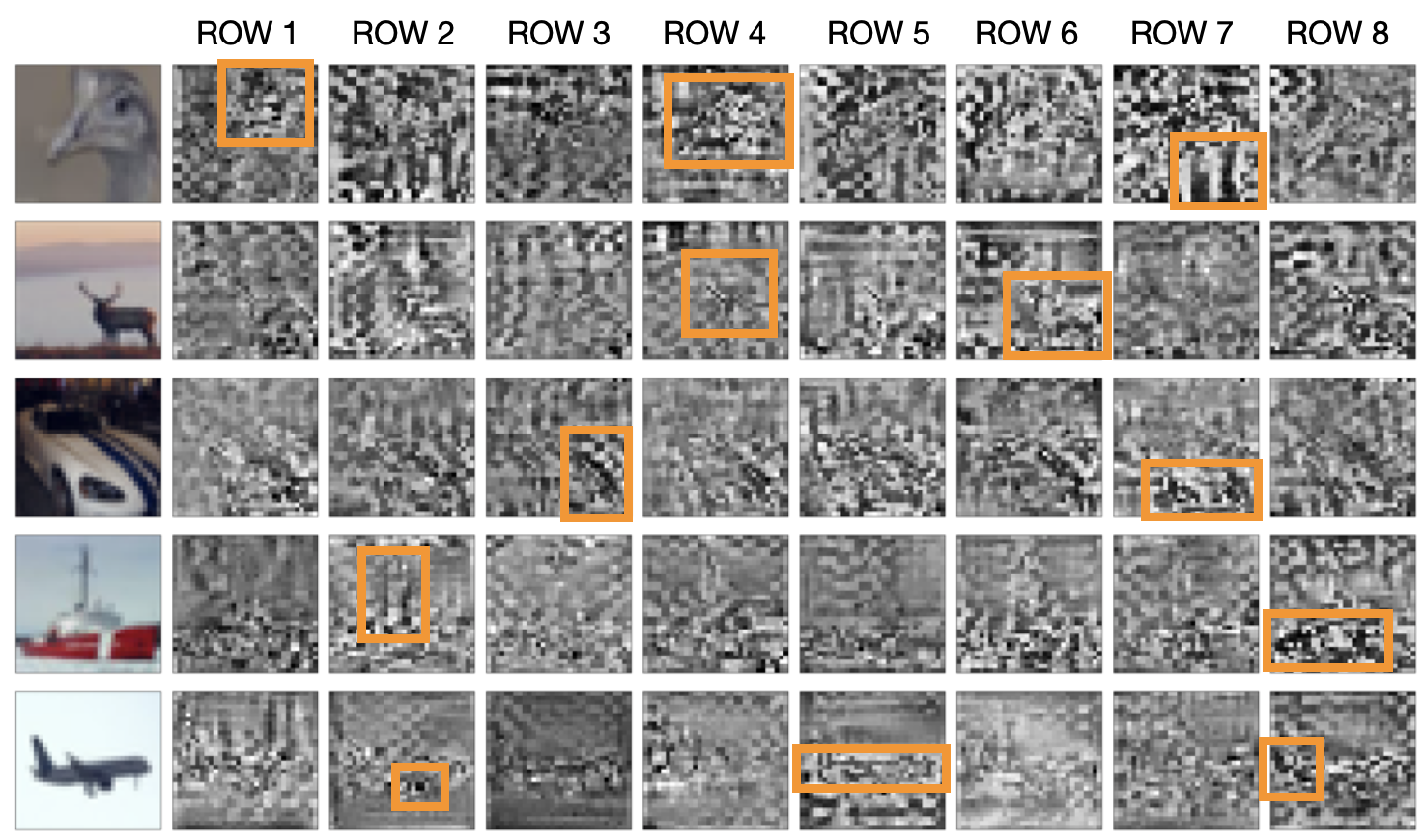}
    \caption{Interpretation of latent representations for different images ($S=8$). Rows tend to individualize on specific portions of the input and in some cases they are semantically meaningful interpretation (as highlighted by orange boxes).}
    \label{fig:grid}
\end{center}
\vskip -0.2in
\end{figure}

\textbf{Interpretation.} Last but not least, we run the proposed interpretation procedure on the latent representations of the Conv-4 backbone for $20000$ iterations and a batch of $64$ samples. Figure~\ref{fig:grid} shows the different masks associated to each row $\pmb{z}_{i,:}$ and each training instance.
We observe that some rows learn to specialize on particular portions of the input, thus carrying semantically meaningful interpretations.

All these experiments provide an initial evidence on the benefits of learning structured representations in an unsupervised way. Indeed, we believe that performance, both in terms of generalization and interpretation, can be improved by introducing other inductive biases. For example, one could replace the global average pooling in the last layer of a backbone with an attentional layer, like slot attention~\cite{locatello2020object}, in order to better cluster the input data and increase the level of specialization of the latent representations. Other possibilities consist of using mutual information bounds different from the one considered in this work. We leave all these analyses to future study.

{\section*{Acknowledgements}\label{sec:acks}}
This research was partially supported by TAILOR, a project funded by EU Horizon 2020 research and innovation programme under GA No 952215.

\bibliography{ref}
\bibliographystyle{icml2021}

\appendix
\section{Simplification of Generalized Mutual Information}

\begin{align}
    \mathcal{I}(X,Z,A) &= E_{p_{X,Z,A}}\Big\{\log\frac{p_{X,Z,A}}{p_X p_Z p_A}\Big\} \nonumber\\
    &= E_{p_{X,Z,A}}\Big\{\log\frac{p_{Z,A|X}}{p_Z p_A}\Big\} \nonumber\\
    &= E_{p_{X,Z,A}}\Big\{\log\frac{p_{Z|X}p_{A|X}}{p_Z p_A}\Big\} \nonumber\\
    &= E_{p_{X,Z,A}}\Big\{\log\frac{p_{Z|X}}{p_Z}\Big\}{+}E_{p_{X,Z,A}}\Big\{\log\frac{p_{A|X}}{p_A}\Big\} \nonumber\\
    &= E_{p_{X,Z}}\Big\{\log\frac{p_{Z|X}}{p_Z}\Big\}{+} E_{p_{X,A}}\Big\{\log\frac{p_{A|X}}{p_A}\Big\} \nonumber\\
    &= E_{p_{X,Z}}\Big\{\log\frac{p_{Z|X}p_X}{p_Z p_X}\Big\}{+} E_{p_{X,A}}\Big\{\log\frac{p_{A|X}p_X}{p_A p_X}\Big\} \nonumber\\
    &=\mathcal{I}(X,Z)+\mathcal{I}(X,A) \nonumber
    \end{align}
    
\section{Derivation of Lower Bound}
We recall the derivation of the lower bound only for the term $\mathcal{I}(X,Z)$, as the other one can be obtained by applying the same methodology.
\begin{align}
    \mathcal{I}(X,Z) &= E_{p_{X,Z}}\Big\{\log\frac{p_{X|Z}}{p_X}\Big\} \nonumber\\
    &= E_{p_{X,Z}}\Big\{\log\frac{p_{X|Z}}{q_{X|Z}}\Big\} + E_{p_{X,Z}}\Big\{\log\frac{q_{X|Z}}{p_X}\Big\} \nonumber\\
    &= E_{p_X}\Big\{KL\{p_{X|Z}\|q_{X|Z}\}\Big\} + E_{p_{X,Z}}\Big\{\log\frac{q_{X|Z}}{p_X}\Big\} \nonumber \\
    &\geq E_{p_{X,Z}}\Big\{\log\frac{q_{X|Z}}{p_X}\Big\}
\end{align}
where $q_{X|Z}$ is an auxiliary density. Now, if we consider $q_{X|Z}=\frac{p_X e^{T_\beta(\pmb{x},\pmb{z})}}{\mathcal{Z}_Z}$, we obtain that:
\begin{align}
    E_{p_{X,Z}}\Big\{\log\frac{q_{X|Z}}{p_X}\Big\} &=
    E_{p_{X,Z}}\Big\{\log\frac{p_X e^{T_\beta(\pmb{x},\pmb{z})}}{p_X \mathcal{Z}_Z}\Big\} \nonumber\\
    &= E_{p_{X,Z}}\Big\{\log\frac{e^{T_\beta(\pmb{x},\pmb{z})}}{\mathcal{Z}_Z}\Big\} \nonumber\\
    &= E_{p_{X,Z}}\{T_\beta(\pmb{x},\pmb{z})\}-E_{p_Z}\{\log\mathcal{Z}_Z\}
\end{align}
By exploiting the inequality $\log x \leq \frac{x}{e}$ for all $x>0$, we obtain that:
\begin{align}
    E_{p_{X,Z}}\Big\{\log\frac{q_{X|Z}}{p_X}\Big\} &\geq
     E_{p_{X,Z}}\{T_\beta(\pmb{x},\pmb{z})\}-\frac{1}{e}E_{p_Z}\{\mathcal{Z}_Z\} \nonumber\\
     &=E_{p_{X,Z}}\{T_\beta(\pmb{x},\pmb{z})\}-\frac{1}{e}E_{p_X p_Z}\Big\{e^{T_\beta(\pmb{x},\pmb{z})}\Big\} \nonumber
\end{align}
Thus concluding the derivation.

\section{Details of Network Architectures}
We use the graph neural network encoder proposed in~\cite{kipf2018neural} and report some details for completeness.

The message passing neural network, i.e. $m_\eta$, is used to predict the relational tensor:
\begin{align}
    \pmb{h}_i^1 &= m_{\text{emb}}(\pmb{z}_{i,:}) \nonumber\\
    \pmb{h}_{(i,j)}^1 &= m_e^1([\pmb{h}_i^1,\pmb{h}_j^1]) \nonumber\\
    \pmb{h}_i^2 &= m_v^1\Big(\sum_{j\neq i}\pmb{h}_{(i,j)}^1\Big) \nonumber\\
    \pmb{h}_{(i,j)}^2 &= m_e^2([\pmb{h}_i^2,\pmb{h}_j^2,\pmb{h}_{(i,j)}^1]) \nonumber\\
    \pmb{a}_{i,j} &= \text{Softmax}\Big((\pmb{h}_{(i,j)}^2+\pmb{g})/\tau\Big) \nonumber
\end{align}
where $m_{\text{emb}}, m_e^1, m_v^1$  and $m_e^2$ are MLPs with one hidden layer and $\pmb{g}\in\mathbb{R}^K$ is a vector whose elements are generated i.i.d from a $\text{Gumbel}(0,1)$ distribution~\cite{maddison2017concrete,jang2017categorical}.

In the experiments on CIFAR-10~\cite{krizhevsky2009learning}, we use $S=D=8$ and $K=2$.

\section{Derivation of Upper Bound}
\begin{align}
 \mathcal{I}(X,Z_{i,:}) &= E_{p_{X,Z_{i,:}}}\Big\{\log\frac{p_{X,Z_{i,:}}}{p_X p_{Z_{i,:}}}\Big\} \nonumber\\
 &= E_{p_{X,Z_{i,:}}}\Big\{\log\frac{ p_{Z_{i,:}|X}}{p_{Z_{i,:}}}\Big\} \nonumber\\
  &= E_{p_{X,Z_{i,:}}}\Big\{\log\frac{q_{Z_{i,:}}p_{Z_{i,:}|X}}{q_{Z_{i,:}}p_{Z_{i,:}}}\Big\} \nonumber\\
  &= E_{p_{X,Z_{i,:}}}\Big\{\log\frac{p_{Z_{i,:}|X}}{q_{Z_{i,:}}}\Big\} - KL\{p_{Z_{i,:}}\|q_{Z_{i,:}}\}\nonumber\\
 &\leq E_{p_{X,Z_{i,:}}}\Big\{\log\frac{p_{Z_{i,:}|X}}{q_{Z_{i,:}}}\Big\} \nonumber\\
 &= E_{p_{X,Z_{i,:}}}\{\log p_{Z_{i,:}|X}\} - E_{p_{Z_{i,:}}}\{\log q_{Z_{i,:}}\} \nonumber\\
 &\propto - E_{p_{Z_{i,:}}}\{\log q_{Z_{i,:}}\} \nonumber\\
 &\propto - E_{p_{Z_{i,:}}}\Big\{\log E_{p_X}\{\mathcal{N}(\tilde{\pmb{z}}_{i,:}|g_{\pmb{\theta}}(\sigma(\pmb{\phi}_i)\odot\pmb{x}),I)\}\Big\} \nonumber\\
 &\leq - E_{p_X p_{Z_{i,:}}}\Big\{\log \mathcal{N}(\tilde{\pmb{z}}_{i,:}|g_{\pmb{\theta}}(\sigma(\pmb{\phi}_i)\odot\pmb{x}),I)\Big\} \nonumber\\
 &\propto E_{p_X p_{Z_{i,:}}}\Big\{\|\tilde{\pmb{z}}_{i,:}-g_{\pmb{\theta}}(\sigma(\pmb{\phi}_i)\odot\pmb{x})\|_2^2\Big\} \nonumber
\end{align}
where the first inequality is obtained by noting that $- KL\{p_{Z_{i,:}}\|q_{Z_{i,:}}\}\leq 0$, and the second one is based on Jensen's inequality.

\section{Hyperparameters in Experiments}
\begin{itemize}
    \item Number of self-supervised training epochs $50$
    \item Batch size $64$
    \item Number of data augmentations $32$
    \item $S=D=8$ and number of relation types $K=2$
    \item Number of supervised training epochs for linear evaluation $100$.
    \item Adam Optimizer with learning rate $1e-3$.
\end{itemize}

\section{Interpretations for the Whole Batch of Samples}
\begin{figure}[h]
\begin{center}
    \includegraphics[width=0.28\columnwidth]{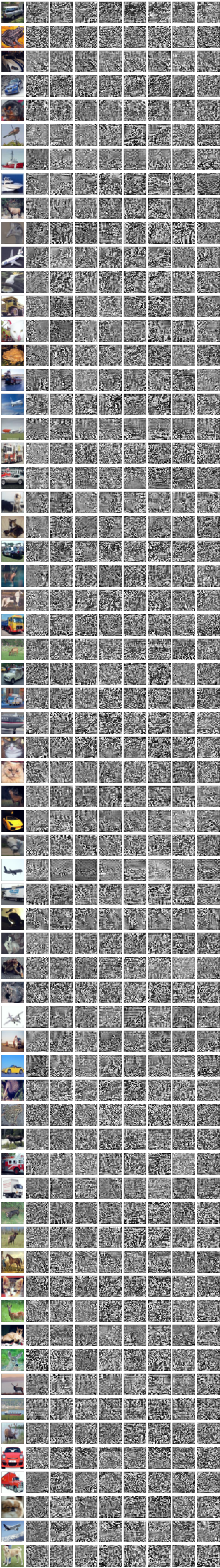}
    \caption{Interpretation of latent representations for different images (better zoom-in).}
    \label{fig:whole_grid}
\end{center}
\vskip -0.2in
\end{figure}

See Figure~\ref{fig:whole_grid}

\end{document}